\pdfoutput=1

\documentclass[11pt]{article}

\usepackage[final]{coling}

\usepackage{times}
\usepackage{latexsym}

\usepackage[T1]{fontenc}

\usepackage[utf8]{inputenc}
\usepackage{microtype}

\usepackage{inconsolata}

\usepackage{graphicx}
\usepackage{amsmath}
\usepackage{cleveref}
\usepackage{makecell}
\usepackage{multirow}
\usepackage{hyperref}

\title{Automated Collection of Evaluation Dataset for Semantic Search in Low-Resource Domain Language}

\author{
 \textbf{Anastasia Zhukova\textsuperscript{1}},
 \textbf{Christian E. Matt\textsuperscript{2}} \and
 \textbf{Bela Gipp\textsuperscript{1}}\\
 \textsuperscript{1}University of G{\"o}ttingen,
 \textsuperscript{2}eschbach GmbH
}

\begin{document}
\maketitle
\begin{abstract}
Domain-specific languages that use a lot of specific terminology often fall into the category of low-resource languages. Collecting test datasets in a narrow domain is time-consuming and requires skilled human resources with domain knowledge and training for the annotation task. This study addresses the challenge of automated collecting test datasets to evaluate semantic search in low-resource domain-specific German language of the process industry. Our approach proposes an end-to-end annotation pipeline for automated query generation to the score reassessment of query-document pairs. To overcome the lack of text encoders trained in the German chemistry domain, we explore a principle of an ensemble of "weak" text encoders trained on common knowledge datasets. We combine individual relevance scores from diverse models to retrieve document candidates and relevance scores generated by an LLM, aiming to achieve consensus on query-document alignment. Evaluation results demonstrate that the ensemble method significantly improves alignment with human-assigned relevance scores, outperforming individual models in both inter-coder agreement and accuracy metrics. These findings suggest that ensemble learning can effectively adapt semantic search systems for specialized, low-resource languages, offering a practical solution to resource limitations in domain-specific contexts.
\end{abstract}

\section{Introduction}

In NLP, a low-resource language lacks sufficient linguistic data, resources, or tools for effective model training and development \cite{hedderich-etal-2021-survey, chu-wang-2018-survey}. Domain-specific German, especially in areas with professional jargon, codes, acronyms, and numeric data, qualifies as a low-resource language because large, publicly accessible datasets for such specialized domains are scarce. As a result, few language models are trained specifically for these areas. While general German has extensive NLP resources, specialized sublanguages often demand unique datasets that are difficult to gather and typically limited in volume.

\begin{figure}
    \centering
    \includegraphics[width=\linewidth]{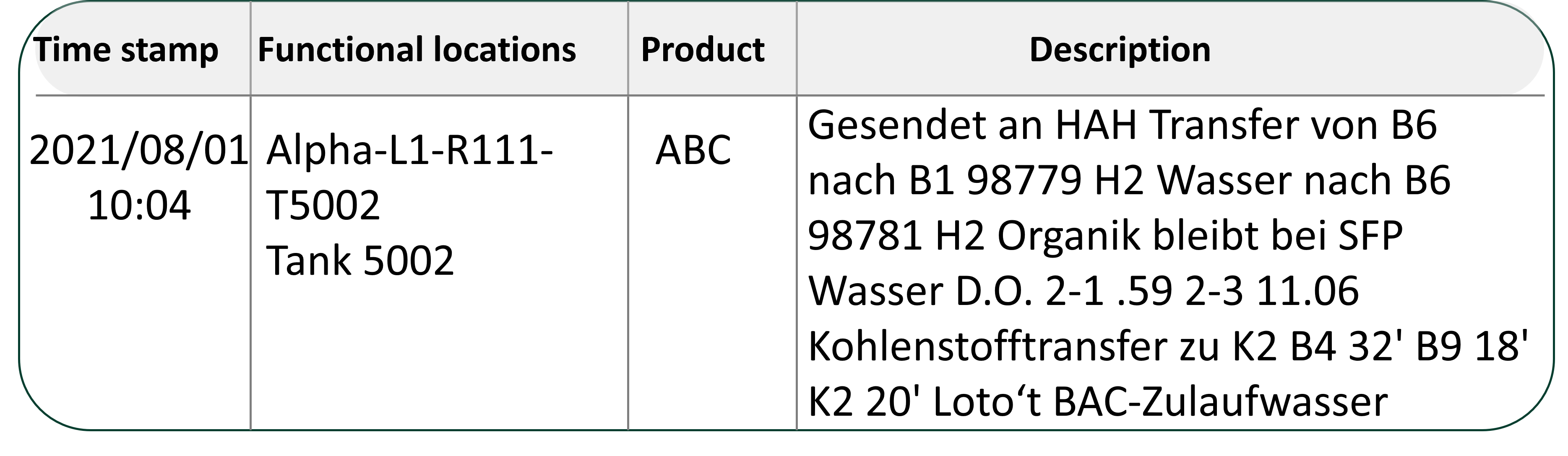}
    \caption{An example of a mocked text log from a shift book in the German language. The logs contain a log of domain-specific terms, which require domain knowledge in the area and know specifics of the production process}
    \label{fig:log_example}
\end{figure}

Shift logs in the process industry are detailed records maintained by operators or technicians during their work shifts (see \Cref{fig:log_example}\footnote{The text in Figure 1 translates to English as ''Sent to HAH Transfer B6 to B1 98779 H2 water to B6 98781 H2 organics still at SFP Water D.O. 2-1 .59 2-3 11.06 Carbon transfer to K2 B4 32' B9 18' K2 20' Loto'd BAC inlet water supply''}). They document key operational activities, system statuses, production metrics, equipment performance, process parameters, maintenance activities, safety observations, product quality, and any incidents or anomalies. The process industry produces and transforms raw materials into finished products through chemical, physical, or biological processes. The complexity of parsing and interpreting professional terminology and industry-specific syntax requires models trained on annotated datasets tailored to the domain, which are often non-existent or proprietary. This lack of accessible, high-quality datasets makes it difficult to build, fine-tune, or adapt existing NLP models for these specialized uses. Without significant efforts in curating and labeling domain-specific data, language models will struggle with accurate interpretation and generation in these fields.

Collecting and annotating text collections for semantic search in low-resource languages presents several significant challenges. First, finding qualified annotators for this task who are both fluent in the language and trained in linguistic annotation can be extremely difficult. Moreover, the complexity of semantic search requires annotations beyond basic syntactic labeling, such as entity recognition and coreference resolution, which demand specialized knowledge and increase the task's difficulty. Second, standalone general language models trained on high-resource languages can collect the test data to a certain extent but do not transfer well to these low-resource contexts and lack accurate language representation of the domain language. 

This paper explores the principle of ensemble learning to create test collections for semantic search in domain-specific German language. Ensemble learning is a machine learning technique that combines multiple individual models, often called "weak learners," to create a more powerful and accurate predictive model by mitigating each other's weaknesses \cite{Ibomoiye-2022}. Our experiments demonstrate that combining an ensemble of multiple encoders with a generative LLM (GPT-4o in our case) to reassess relevance scores significantly improves the quality of test collections for semantic search evaluation. Specifically, this approach increases inter-coder agreement (measured by Krippendorff's alpha) by nearly four times and improves the F1-score by 1.5 times.

\section{Related work}
Ensemble learning improves machine learning performance by combining predictions from multiple models, thus enhancing accuracy, reducing variance, and mitigating bias \cite{Ibomoiye-2022}. Ensemble learning is popular across domain-specific domains and applications, such as medical diagnosis and fraud detection. It has started evolving from being used with machine learning algorithms to deep learning models. 

LLMs have already been widely used for data annotation, specifically for domain-specific tasks requiring specialized domain knowledge, where human annotations are costly but crucial \cite{TanLWB24}. Multiple studies have evaluated LLMs in biomedicine \cite{ZhuZHH23, KumarAYL24}, law and education \cite{ZhuZHH23}, and financial sector \cite{aguda-etal-2024-large}. While LLMs are a powerful tool for data annotation, the studies show that standalone LLMs perform worse than human annotators \cite{LuYZW23, StaffSIP23}. 

To mitigate the drawbacks of LLM annotations, new methods were proposed to involve reasoning, reevaluating the assigned labels, or involving collective decisions. One of the state-of-the-art techniques is to use a human-in-loop annotation process and help human annotators by augmenting them with the fast LLM-pre-annotated labels \cite{li-etal-2023-coannotating}. The most recent development employs an ensemble of LLMs for annotation \cite{FarrMCS24} or utilizes a synergy of thoughts across multiple smaller-scale LMs \cite{ShangLXL24}, similar to ensemble learning with "weak" models. 

\begin{figure*}[h]
  \includegraphics[width=\textwidth]{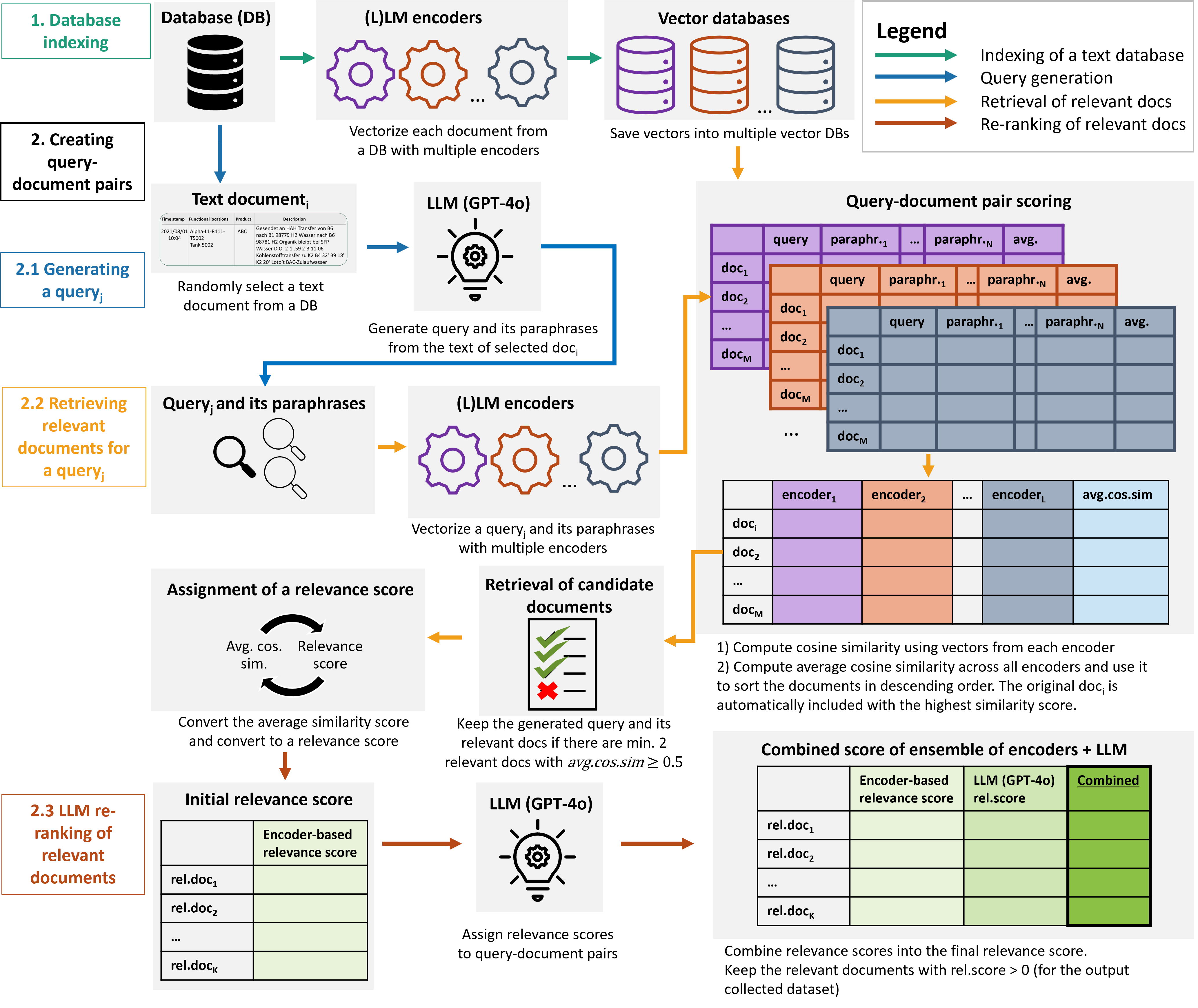}
  \caption{A proposed methodology with ensembles of (L)LM encoders used to retrieve the most relevant documents, i.e., text logs of a shift book, and with an LLM to adjust the relevance score for the document re-ranking.}
  \label{fig:methodology}
\end{figure*}

\section{Methodology}
Ensemble learning is widely used in practice because it can improve model robustness and accuracy and reduce variance, especially when individual models are prone to errors or have high variability. The central idea is that by aggregating the predictions of several models, the ensemble can outperform any single model, reducing the risk of overfitting and improving generalization. Ensemble methods leverage the strengths of different models while compensating for their weaknesses, leading to better performance on complex tasks \cite{Ibomoiye-2022}. In stacking of ensemble learning, different models (often of different types) are trained, and their predictions are used as input to a "meta-model," which learns how to combine these predictions to make the final decision.

The methodology of the ensemble for annotating a test collection for semantic search comprises two main parts: (1) document indexing and (2) creation of the query-document pairs. The key aspect of document indexing is using a set of encoders with various architectures and training strategies. The goal is to combine different aspects of the document similarity that each encoder has learned. Re-ranking combines the relevance score based on the document similarity with the score generated by a generative LLM. LLM assesses the relevance of the query-document pair independently from the score used for the retrieval, thus allowing the combining of another "point of view" to the query-document relevance. \Cref{fig:methodology} depicts the proposed methodology.

\subsection{Database indexing}
Multiple encoders are used for the database indexing. Possible ways to encode a text document include document encoding by the model architecture (i.e., bi-encoder) and mean pooling of the word vectors (additionally, see \Cref{sec:discussion}). Each document encoder may have learned different vector representation from the others due to its architecture, training setup, and dataset on which it was trained. We encode with multiple encoders to use this diversity of the vector representation. 

For our experiments, we used three bi-encoders: two based on the sentence transformer architecture and one text encoder from OpenAI\footnote{In our implementation we used \href{https://learn.microsoft.com/en-us/azure/ai-services/openai/concepts/models?tabs=python-secure\#embeddings-models}{azure-text-embedding-3-large} with a private endpoint.}. We selected the models that supported German, had a strong performance on the semantic search on the publicly available benchmarks, yielded the best results on a small manually created dataset (see \Cref{appendix:models} for more details), and could use cosine similarity as a score metric (see \Cref{sec:query_gen}). Each document, i.e., a text log,  is of a size between a sentence and paragraph and was encoded based on the input capacity of an encoder, i.e., truncated if needed. 

\subsection{Creating query-document pairs}
\label{sec:query_gen}
\paragraph{Query generation}
A query was generated from a randomly selected document, i.e., a text log from a database, to ensure that at least one document was relevant to a query. We chose only long enough documents for the query generation, i.e., at least 100 chars. A query was generated with an LLM; in our implementation, it was GPT-4o. The prompt was designed to make generated queries extracted keywords from the text that look like search queries. Following the principle of rewriting a search query in real life to retrieve more fitting documents, the same prompt additionally generated paraphrases to the query: 

{\small
\begin{verbatim}
Extract {query_num} search queries from the
following text \'{text}\'. The queries need to be 
meaningful as if you are supposed to use them to 
google. A query should contain between 2 to 5 
words. Minimize using tokens with digits. Avoid 
using persons' names. Paraphrase each extracted 
query into 2 to 4 modifications. When creating 
paraphrases, make them look like you want to 
reformulate them for better search results. The 
paraphrases should contain synonyms of the 
original words in a query or syntactically 
correct change of the word order. Reply with a 
list of strings, with each string a query 
followed by its modifications separated by a 
semicolon. Keep only the text of queries, no 
enumeration. Consider the entire context, as it 
is crucial for understanding the text. The texts
are from the context of chemical and 
pharmaceutical production environments.
\end{verbatim}
}

If a document was long enough (i.e., more than 300 chars), multiple queries were generated and used in the annotation pipeline. We tracked a list of the documents already used in the query generation and kept selecting only the unused ones. 

\paragraph{Retrieval}
We used the linear search on the L2-normalized vectors with cosine similarity as a similarity score function. We did not use other techniques to ensure each document would acquire a similarity score. We followed a two-step approach to make the final similarity score used for the retrieval more robust. 

First, we computed a similarity score independently between a query and all documents and the query's paraphrases and all documents. Using paraphrases enables retrieval of a more complete list of documents than solely using the original query by covering a wider lexical diversity used in the text. The mean score per encoder is used as an intermediate similarity score for a document $d$ and query $q$ vectorized by encoder $e_l$: 
\begin{equation}
cos.sim_{d, e} = \frac{1}{|QP|}\sum_{q \in QP}^{} cos.sim_{q, (d, e)}
\end{equation}
where $QP$ is a set of a query and its paraphrases, and $|QP|$ is the size of this set. 

Second, we average the scores across all encoders, thus scoring query-document similarity equally by all used vector models: 
\begin{equation}
cos.sim_{d} = \frac{1}{|E|}\sum_{e \in E}^{} cos.sim_{d, e}  
\end{equation}
where $E$ is a set of the used encoders, and $|E|$ is the size of this set. 

Despite the calculated score, the similarity score of the original document is assigned to 1.0 to ensure that it will be among the retrieved documents and has the highest score. 

Lastly, we retrieve the best-matching documents and assign relevance scores to the query-document pairs. To decide which documents to retrieve, we check two conditions: (1) the documents must have $cos.sim_{d} \geq 0.5$, and (2) per query, should have at least two relevant documents. The following function converted the cosine similarity to the relevance score of the ensemble of encoders on a scale of 1 to 3, where 3 meant high relevance of a document to a query, 2 was partial relevance, and 1 referred to marginal relevance:

\begin{equation}
ensemble_{d} = 
    \begin{cases}
    1 & \text{if } 0.5 \leq cos.sim_d < 0.6, \\
    2 & \text{if } 0.6 \leq cos.sim_d < 0.7, \\
    3 & \text{if }  cos.sim_d \geq 0.7\
    \end{cases}
\end{equation}

\paragraph{Re-ranking}
The goal of re-ranking was to use an LLM to assess the query-document pair independently from the encoders, and (1) use its relevance score combined with the encoders' score, (2) check if an LLM reevaluated the pairs as irrelevant, i.e., assigned 0 scores:

{\small
\begin{verbatim}
Assign a relevance score between 3 
to 0 of how a query \'{query}\' matches an 
event \'{text}\' which occurred at a machinery 
\'{funcloc}\'. 3 is a strong relevance, i.e., 
a document directly contains the information 
requested in a query. The relevance is 
strong if the query matches a document on a
synonym level and some spelling modifications
(including a match of a full phrase/word to 
its abbreviation/shortening). 2 is a middle 
relevance, i.e., a document contains only 
some terms or synonyms (more than 1) or the 
information in a document refers to an adjacent 
element in a text. For example, a query 
specifies a specific type of container that 
is empty, and a document contains a different 
type of container that is empty. Score 1 means
little relevance, i.e., a document partially 
contains some information requested in a query,
e.g., some terms from the query but 
distributed across the document or only 1-2 
terms/synonyms from a query are mentioned in a
document, but they don’t belong to one 
neighborhood to reflect the semantics of a 
query. For example, for a query \'pump is 
defective\' some document contains general 
information about a pump. A score of 0 means 
that a document is not relevant to a query. 
Output only the relevance score in an integer
between 0 and 3.
\end{verbatim}
}

Combining the relevance scores from the two sources, i.e., encoders and LLM, is done with the following formula:

\begin{equation}
\small
combined_d = 
    \begin{cases}
    0, \text{if } LLM_d = 0, \\
    bins(\frac{2 \cdot LLM_d + ens_d}{3}), \text{if } LLM_d = 3, \\
    bins(\frac{LLM_d + 2 \cdot ens_d}{3}), \text{if } ens_d = 1, \\
    bins(\frac{LLM_d + ens_d}{2}), \text{else}   \\
    \end{cases}
\end{equation}
where 
\begin{equation}
bins(x) = 
    \begin{cases}
    3 & \text{if } x \geq 2.6, \\
    2 & \text{if } 2.0 \leq x < 2.6,  \\
    1 & \text{if } 1.0 \leq x < 2.0  \\
    0 & \text{if }  x \leq 1.0\\
    \end{cases}
\end{equation}

\begin{figure}[h!]
    \centering
    \includegraphics[width=0.6\linewidth]{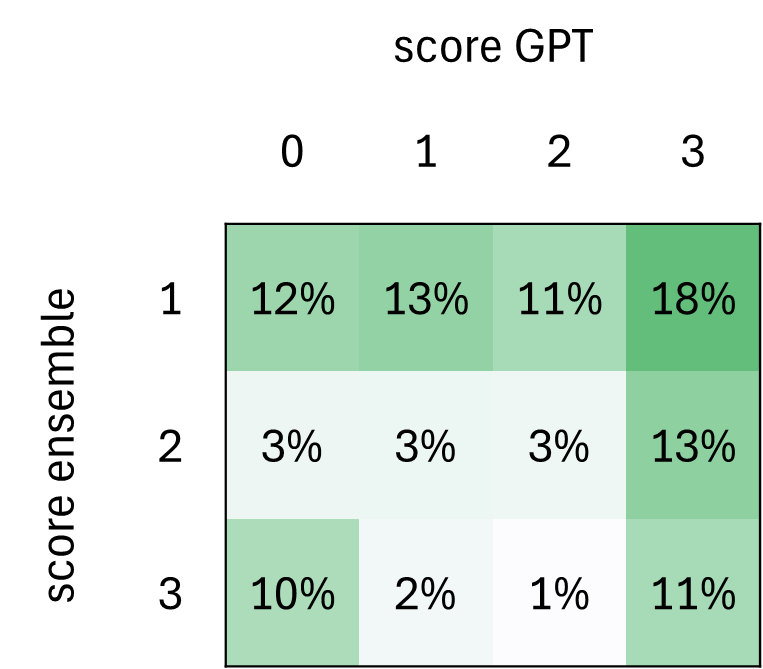}
    \caption{The distribution of the relevance scores produced by an ensemble of encoders and GPT-4o. While the ensemble assigns 1 relevance score, GPT-4o leans towards the score of 3. The proposed combined approach balances out these model tendencies. }
    \label{fig:ensemble_vs_gpt}
\end{figure}

The formula for the \textit{combined} relevance score originates from the moderate agreement between the ensemble and LLM. \Cref{fig:ensemble_vs_gpt} shows that most scores were either annotated by the ensemble as 1 or by the GPT-4o as 3. Hence, when computing the combined score, we give more weight to the GPT scores when the score is 3 or to the ensemble scores when it is 1; otherwise, we compute their average. Moreover, GPT-4o tends to re-rank the fourth of the ensemble-positive scores as 0. Therefore, we keep the re-ranking score of GPT-4o. The $bins(x)$ function was empirically derived from our experiments. 

\section{Evaluation}
We evaluated our approach against the manually assigned relevance scores to the retrieved documents \cite{PangakisWF23}. The goal was to evaluate how the proposed approach agreed with how a human assessed the query-document pairs.

\begin{figure*}[h]
\centering
  \includegraphics[width=\textwidth]{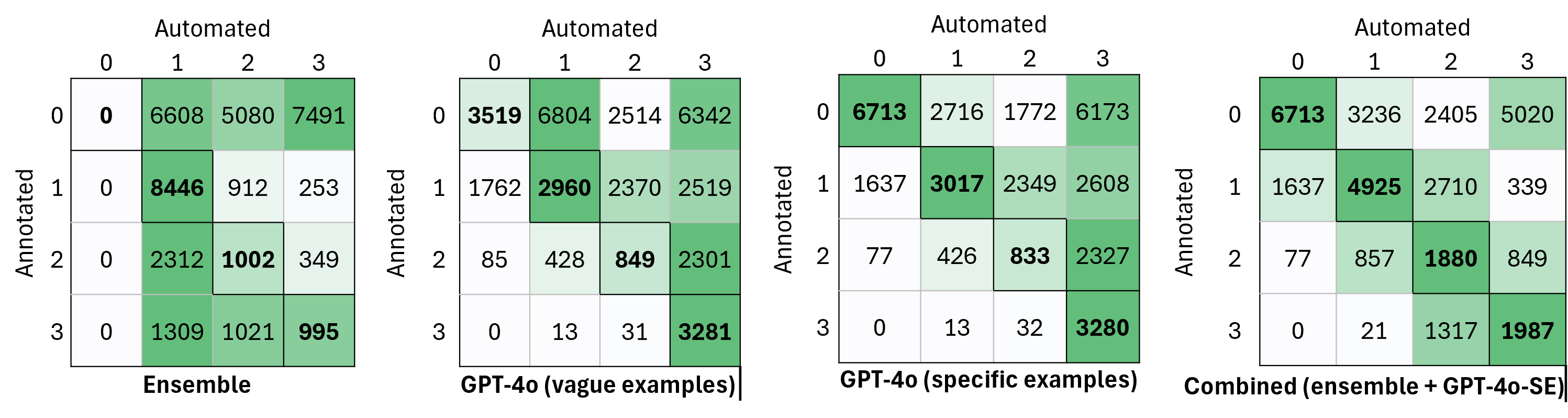}
  \caption{The confusion matrices of the annotated vs. automated relevance scores for four methods: an ensemble of encoders, GPT-4o with vague examples, GPT-4o with specific examples (SE), and combined ensemble + GPT-4o-SE. The combined approach allocates most of the results on the matrix diagonal, whereas its components separately lean towards one score or another.}
  \label{fig:confusion_matrices}
\end{figure*}

\subsection{Experiments}
We used the approach to create a test collection from seven plant shift books. We have generated at least 80 queries for each source for which at least two relevant documents were identified.  We selected 28-30 queries with up to 1000 relevant documents each for the manual annotation to make the task feasible. We provided a native German speaker familiar with the domain, and the instructions were identical to those used in the prompt. The documents were already sorted by the automated relevance scores, but the hired annotator was to assign the relevance scores between 3 and 0 without seeing these scores. Since recall-based evaluation is impossible, i.e., evaluating how many documents were retrieved from the overall number of relevant documents, we focus on evaluating final relevance scores.

\begin{table*}[h!]
\small
\centering
\begin{tabular}{c|c|c|r|r|r|r|r}
\hline
\textbf{Source} & \textbf{Stats} & \textbf{Model} & \textbf{\makecell[r]{Kripp.'s \\alpha}} & \textbf{Precision} & \textbf{Recall} & \textbf{F1} & \textbf{nDCG} \\ 
\hline
\multirow{3}{*}{\textbf{A} } & All docs: 17053 & Ensemble & 50.30 & 38.24 & 39.66 & 33.61 &  \textbf{97.71} \\ 
                            & \# queries: 30 & GPT-4o-VE  & 31.49 & 52.29 & 41.55 & 38.63 &  95.37 \\ 
        & \# verified retrieved candidates: & GPT-4o-SE & 44.10               & 56.51    & 46.28  & 44.39     & 95.60  \\
        & 2739 & Combined     & \textbf{67.03}               & \textbf{60.90}    & \textbf{53.42}  & \textbf{54.89}     & 97.60  \\
\hline
\multirow{3}{*}{\textbf{B}} & All docs: 14065 & Ensemble & 55.57 & 45.49 & 42.02 & 39.97 &  98.01 \\ 
                            & \# queries: 30 & GPT-4o-VE  & 40.37 & 43.28 & 42.14 & 38.45 & 95.32 \\ 
        & \# verified retrieved candidates: & GPT-4o-SE & 45.61               & 45.07    & 44.34  & 40.79    & 95.60  \\
        & 2022 & Combined     & \textbf{68.69}               & \textbf{51.72}    & \textbf{49.50}  & \textbf{49.96}    & \textbf{98.05}  \\
\hline
\multirow{3}{*}{\textbf{C}} & All docs: 129345 & Ensemble & 31.55 & 36.33 & 35.20 & 32.15  & 93.49 \\ 
                            & \# queries: 30 & GPT-4o-VE  & 44.41 & 41.17 & 53.40 & 37.47 & 93.62 \\ 
        & \# verified retrieved candidates: & GPT-4o-SE & 46.70               & 41.37    & \textbf{54.18}  & 37.89     & 93.74  \\
        & 2166 & Combined     & \textbf{61.35}               & \textbf{46.92}    & 51.33  & \textbf{48.16}   & \textbf{95.35}  \\
\hline
\multirow{3}{*}{\textbf{D}} & All docs: 70823 & Ensemble & 14.39 & 18.56 & 25.81 & 21.25  & 90.44 \\  
                            & \# queries: 30 & GPT-4o-VE  & 31.40 & 45.65 & 49.42 & 39.85 & 93.11 \\ 
        & \# verified retrieved candidates: & GPT-4o-SE & 38.11               & 45.64    & \textbf{50.48 } & 41.11     & 93.72  \\
        & 5111 & Combined     & \textbf{54.71}               & \textbf{50.87}    & 48.47  & \textbf{46.38 }    & \textbf{94.64}  \\
\hline
\multirow{3}{*}{\textbf{E}} & All docs: 9730 & Ensemble & \textbf{81.60} & 8.34 & 36.16 & 12.73  & 59.09 \\ 
                            & \# queries: 28 & GPT-4o-VE  & -39.27 & 24.07 & 39.85 & 11.32  & 67.64 \\ 
       & \# verified retrieved candidates: & GPT-4o-SE & -23.67              & 27.56    & 41.68  & 20.29    & \textbf{68.52}  \\
       & 7562 & Combined     & -24.91              & \textbf{30.00}    & \textbf{48.40}  & \textbf{24.19}     & 66.05  \\   
\hline
\multirow{3}{*}{\textbf{F}} & All docs: 25752 & Ensemble & 8.56 & 31.16 & 33.42 & 28.01  & 89.88 \\ 
                            & \# queries: 28 & GPT-4o-VE  & 26.33 & 39.60 & 44.14 & 36.66 &  91.16 \\ 
        & \# verified retrieved candidates: & GPT-4o-SE & 39.97               & 45.95    & \textbf{49.05}  & 42.82     & \textbf{92.72}  \\
        & 2741 & Combined     & \textbf{41.74}               & \textbf{46.48}    & 46.79  & \textbf{44.79}     & 91.63  \\
\hline
\multirow{3}{*}{\textbf{G}} & All docs: 63570 & Ensemble & -2.31 & 23.53 & 38.09 & 28.06 & \textbf{87.57} \\ 
                            & \# queries: 29  & GPT-4o-VE  & -3.68 & 26.62 & 42.49 & 21.71  & 86.22 \\ 
        & \# verified retrieved candidates: & GPT-4o-SE & 1.12                & 32.69    & \textbf{45.55}  & 25.97     & 86.78  \\
        & 4406 & Combined     & \textbf{14.90}               & \textbf{34.19 }   & 39.44  & \textbf{30.35}     & 86.65  \\
\hline
\multirow{3}{*}{\textbf{Average}} & All docs: 330338 & Ensemble & 10.92 & 28.81 & 35.77 & 27.97  & 88.03 \\ 
                            & \# queries: 205 & GPT-4o-VE  & 18.72 & 38.96 & 44.71 & 32.01 &  88.92 \\ 
        & \# verified retrieved candidates: & GPT-4o-SE & 27.42               & 42.11    & 47.37  & 36.18     & 89.52  \\
        & 26747 & Combined     & \textbf{40.50}               & \textbf{45.87}    & \textbf{48.19}  & \textbf{42.68}    & \textbf{90.00} \\
\hline

\end{tabular}
\caption{The proposed approach of combining relevance scores produced by an ensemble of text encoders and reranking by GPT-4o yields, on average, the best results in three types of metrics, i.e., intercoder agreement, accuracy, and ranking. }
\label{tab:results}
\end{table*}

\paragraph{Metrics}
We selected a set of diverse metrics to evaluate the automated assignment of the relevance score defined as various tasks: (1) inter-coder agreement between two annotators (i.e., automated and manual) measured by Krippendoff's alpha, (2) classification metrics for the imbalanced classes, such as macro precision, recall, and F1-score, (3) a ranking metric for information retrieval and recommender systems, such as nDCG.  

\textbf{Krippendorff's alpha} is a robust statistical measure utilized to evaluate the reliability or inter-rater agreement across multiple annotators in categorizing or labeling data \cite{krippendorff2013content}. Unlike other agreement metrics, Krippendorff's alpha is versatile, accommodating different levels of measurement, including nominal, ordinal, interval, and ratio scales. The metric yields a value between 0 and 1, where 1 signifies perfect agreement, and 0 indicates no agreement beyond chance. Due to its adaptability and rigorous assessment of inter-rater reliability, Krippendorff's alpha is extensively employed in fields such as content analysis and qualitative data coding, where ensuring the consistency of human judgment is critical.

In the context of imbalanced datasets, \textbf{macro-averaged precision, recall, and F1-score} provide a more balanced evaluation of classification models by giving equal weight to each class, regardless of its frequency. Macro precision, recall, or F1-score first calculates these metrics for each class. Then, it averages the results, ensuring that smaller minority classes are not overshadowed by the majority class and helping to assess the model's ability to avoid false positives across all classes. This approach is particularly useful for imbalanced datasets, where traditional accuracy measures might be skewed by the model's performance on the dominant class. At the same time, macro-averaging ensures a fair evaluation of all classes.

\textbf{Balanced accuracy} is a metric designed to evaluate classification performance on imbalanced datasets, where traditional accuracy may be misleading due to the disproportionate representation of classes. It is calculated as the average of the true positive rate (recall) for each class, ensuring that all classes, including the minority class, are equally considered. Unlike standard accuracy, which can be inflated by the model’s performance on the dominant class, balanced accuracy provides a more equitable assessment by giving equal weight to both the positive and negative classes, regardless of their prevalence in the dataset. This makes it a more robust metric for evaluating models in scenarios where class imbalance is a concern, as it reflects the model’s ability to classify both frequent and infrequent classes correctly.

Normalized Discounted Cumulative Gain (\textbf{nDCG}) is a widely used evaluation metric for ranking tasks, particularly in information retrieval and recommender systems \cite{Liu09}. It measures the ranking quality by comparing the predicted order of items to the ideal, or ground truth, ranking. nDCG is based on the Discounted Cumulative Gain (DCG) concept, which assigns higher relevance scores to items ranked at the top of the list by applying a logarithmic discount factor to lower-ranked items. This emphasizes the importance of correctly ranking more relevant items higher. nDCG normalizes this score by dividing the DCG by the ideal DCG (IDCG)—the DCG of the perfect ranking—resulting in a value between 0 and 1. A score of 1 indicates a perfect ranking, while lower scores reflect the degradation in ranking quality. This metric is particularly useful in scenarios where the relevance of items decreases with their position in the ranked list, making it a robust measure for evaluating the effectiveness of ranked outputs.

\paragraph{Baselines}
To measure the impact of each of these components within the proposed approach, we compare the proposed approach to ranking solely with the ensemble of encoders (Ens.) or GPT-4o (GPT). Moreover, we compare GPT-4o scores produced by two versions of prompts: with vaguely formulated examples of query-document relevance (\textit{GPT-4o-VE}) and specific examples (\textit{GPT-4o-SE}) of the pairs and corresponding scores\footnote{We report here only a prompt with vague examples of what we used in our experiments. We cannot provide prompts with specific examples because they contain proprietary data.}. The proposed approach is denoted as \textit{Comb.} and consists of a combined ensemble of encoders and GPT-4o re-ranking prompted with specific examples.

\paragraph{Results}
\Cref{tab:results} reports metrics computed per method across 7 created test collections and their average. The table shows that the proposed method of combining an ensemble of encoders and GPT-4o outperformed these methods applied independently. The approach outperformed the baselines in all metrics, but Krippendorff's alpha measures the most significant impact. Combining the relevance scores produced by an ensemble of encoders with GPT-4o, on average, improved the inter-coder agreement by a factor of 4. The results also show that providing explicit examples of query-document pairs with their corresponding scores systematically improves all metrics compared to a prompt with vague examples.

Further, we built confusion matrices to see how the score assignment was distributed between manually annotated and automated relevance scores. \Cref{fig:confusion_matrices} shows that the annotator often assessed the query-document pairs as irrelevant despite the score. Moreover, we see that the ensemble of encoders assigned a lot of pairs to score 1, whereas GPT-4o tends to assess the pairs more positively, with a score of 3 in many cases. Providing examples of query-document pairs with positive relevance scores has improved the correct assignment of the 0 score. \Cref{tab:recall} shows recall computed based on these matrices. The ensemble of encoders has the highest recall score of 1, whereas all versions of GPT-4o have the highest recall score of 3. Combining both yields the highest result on score 2 (which seems to be the hardest category to decide) and the highest average recall.

\begin{table}[]
\small
\begin{tabular}{l|l|l|l|l}
\hline
Rel.score    & Ens. & GPT-4o-VE & GPT-4o-SE & Comb. \\
\hline
0   & --    & 18.3  & 38.6 &	38.6    \\
1   & \textbf{87.9}    & 30.8  & 31.4	& 51.2    \\
2   & 27.4    & 23.2  & 22.7	& \textbf{51.3}    \\
3   & 29.9    & \textbf{98.7}  & 98.6	& 59.8    \\
\hline
average & 36.3    & 42.8  & 47.9	& \textbf{50.2}   \\
\hline
\end{tabular}
\caption{Recall the score classification compared to the manually assigned relevance scores. Providing specific examples on prompting (GPT-4o-SE) outperformed prompting with vague examples (GPT-4o-VE), with the most noticeable improvement in recognizing irrelevant query-document pairs, which scored as 0. Combining an ensemble of encoders (Ens.) with GPT-4o-SE yielded worse recall for relevance scores 1 and 3 but significantly  improved the recall on the more ambiguous score 2. }
\label{tab:recall}
\end{table}

\subsection{Discussion and future work}
\label{sec:discussion}

The evaluation results show that combining multiple relevance scores from diverse scoring methods increases the approach's agreement and performance. We tested the approach on the low-resource language of the domain-specific German used on the production sights. Although the approach reaches moderate agreement with the human labels, it can produce a large-scale, diverse evaluation collection with minimum human annotation effort. If the final relevance scores are not ideal and still require manual verification of the query-document pairs, the time required for it is considerably lower than performing the full annotation pipeline from scratch. Below, we discuss the findings, possible adjustments to the other languages, and further improvements.

\paragraph{Zero- vs few-shot learning for the domain-specific tasks} Our experiments have shown that providing specific examples of the query-document pairs and describing how to assign each score enables LLMs to provide more accurate scores. These examples in the few-shot learning setup help shift an LLM towards a domain of interest, which is crucial in prompting an LLM mainly trained on the data with common knowledge towards a specific knowledge area.

\paragraph{Other languages}
Nowadays, there is a vast majority of publicly available and commercial document encoders\footnote{Some examples of commercial encoders are \href{https://platform.openai.com/docs/guides/embeddings/embedding-models}{OpenAI embeddings} and \href{https://cohere.com/embed}{Cohere}}. For example, some sentence transformer models support 50 languages\footnote{\url{https://huggingface.co/sentence-transformers/paraphrase-multilingual-MiniLM-L12-v2}}. A model store of HuggingFace comes in handy for selecting suitable document encoders for an ensemble of encoders. One of the most recent public multilingual encoders is E5 Text Embeddings \footnote{\url{https://huggingface.co/intfloat/multilingual-e5-base}} \cite{wang2024multilingual} trained for 94 languages.  Another hub of a vast selection of encoders is available via LangChain integration\footnote{LangChain supports \href{https://python.langchain.com/docs/integrations/text\_embedding/}{official integration of embeddings or APIs} and offers  \href{https://python.langchain.com/api\_reference/community/embeddings.html}{community API} for more models  }. Moreover, for sentences or short paragraphs, mean pooling of the word vectors can serve as an extra document encoding method. For example, fastText supports 157 languages\footnote{\url{https://fasttext.cc/docs/en/crawl-vectors.html}} and has already been applied as a document encoder for a domain-specific language \cite{ZhukovaHG21, Zhukova2024}. 

The recent releases of multiple public multilingual LLMs make the methodology more feasible to expand to more languages. For example, LlaMa 3\footnote{\url{https://ai.meta.com/blog/meta-llama-3/}}, EuroLLM-9B\footnote{\url{https://huggingface.co/utter-project/EuroLLM-9B}}, Salamandra-7B\footnote{\url{https://huggingface.co/BSC-LT/salamandra-7b}}, and OpenGPT-X Teuken-7B\footnote{\url{https://huggingface.co/openGPT-X/Teuken-7B-instruct-research-v0.4}} can be used instead of GPT-4o for query generation and re-ranking query-document pairs as a free alternative. Still, the performance comparison of these models compared to GPT-4o remains for further investigation. 

\paragraph{Further improvements}  Despite the approach performing better than the baselines, the final metrics can be interpreted as weak agreement or moderate effectiveness. The proposed approach of combining the similarity scores and, later, the relevance scores is rather naive and can be improved. First, the encoders may have a more sophisticated way of score combination, e.g., from reliability weight per encoder score to the loss function that will minimize disagreement between the encoders. Second, multi-agent LLMs can be used to solve a complicated task of the query-document relevance assessment \cite{SuzgunK24, Becker24, Yang_Dailisan_Korecki_Hausladen_Helbing_2024}, or alternatively, various LLMs can be asked to perform the same task \cite{yin-etal-2023-exchange, TanLWB24}.   

\section{Conclusion}
This paper investigates a principle of ensemble learning with "weak" text encoders to create a test collection for the semantic search evaluation. We combined multiple text encoder models for document retrieval. We experimented with creating a test collection for semantic search evaluation in the domain of the German process industry. The experiments showed that computing the final relevance score by combining the average score of the ensemble of text encoders and an independent relevance score created by an LLM for each query-document pair increases the inter-coder agreement and accuracy metrics for several datasets. We invite the research community to apply further and investigate the proposed methodology across additional languages and domains.

\section{Limitations}
The methodology for automated data collection for semantic search in low-resource languages faces several limitations.

\paragraph{Limited Access to Commercial LLMs} The lack of accessibility to commercial APIs of LLMs can lead to different results when relying on publicly available LLMs than those reported. These public models may not have the same performance or language support as commercial offerings, making it difficult to ensure reliable and high-quality data collection across different low-resource languages. 

\paragraph{Ethical and Legal Constraints using LLMs} Depending on a domain, using public APIs or publicly hosted LLMs, e.g., on a university cluster, may not be possible. For instance, the legal constraints around data privacy in the healthcare domain (e.g., GDPR compliance) may be stricter than in other industries, necessitating different data handling practices. This could limit the generalizability of the methodology when crossing into different regulatory environments.

\paragraph{Different prompting requirements}  Low-resource languages may require tailored prompting strategies to extract meaningful and accurate data from LLMs. A prompting approach that works for one language or model might not generalize well to others, necessitating the design of custom prompts for each language or LLM, adding complexity to the automated data collection process.

\paragraph{Lack of multiple strong text encoders} Not all low-resource languages have sufficient encoder-based language models for effective use in automated data collection. Some languages may have only one or even no pre-trained encoders, limiting the ability to implement encoder-decoder architectures commonly used in semantic search, which could reduce performance and accuracy for these languages.

\paragraph{Complex adjustments for other downstream tasks} Automated collection of datasets for downstream tasks, such as named entity recognition, sentiment analysis, or machine translation, may require significant adjustments for low-resource languages. This could involve re-tuning models, modifying preprocessing pipelines, or adapting annotations, which can be time-consuming and resource-intensive, hindering the scalability of the methodology across different languages.

\section{Ethic considerations}
\paragraph{Data Privacy and Consent}  The sensitive private data used in these studies is protected under GDPR regulations, ensuring full compliance with privacy laws. As a result, explicit consent from data subjects was obtained where required. Due to GDPR restrictions, specific examples or direct details regarding the data cannot be provided. Additionally, anonymization techniques were applied to safeguard personal information.

\paragraph{Transparency and Accountability}  The code, datasets, and implementation details that can be shared publicly have been fully discussed, with links provided throughout the main paper and appendix. These resources ensure the research is transparent and can be replicated and scrutinized. However, parts of the work that fall under commercial secrets cannot be revealed due to proprietary restrictions. This limitation impacts transparency, but necessary steps have been taken to share as much as possible without violating commercial confidentiality.

\section*{Acknowledgments}
This Project is supported by the Federal Ministry for Economic Affairs and Climate Action (BMWK) on the basis of a decision by the German Bundestag.

\bibliography{custom}

\appendix
\section{Appendix}

\subsection{Selection of the bi-encoder models}
\label{appendix:models}

The following section describes the methodology of the semi-automated collection of the test dataset for semantic search. The produced dataset is an intermediate version that helped navigate the decision on the model selection for the ensemble of encoders. 

We selected five publicly available text encoders and one commercial, all supporting the German language. All models use cosine similarity as a similarity metric. 

\begin{table*}[h!]
\small
\centering
\begin{tabular}{l|r|r|r|r|r|r|r}
\hline
\textbf{Models} &\textbf{ P@10} & \textbf{R@10} & \textbf{F1@10} & \textbf{MAP@10} & \textbf{MRR} & \textbf{nDCG@10} & \textbf{AVG} \\
\hline
\href{https://huggingface.co/T-Systems-onsite/german-roberta-sentence-transformer-v2}{\makecell[l]{T-Systems-onsite/\\german-roberta-sentence-transformer-v2}} & 16.84 & 9.08 & 9.85 & 38.02 & 45.49 & 20.25 & 23.26 \\
\hline
\href{https://huggingface.co/PM-AI/bi-encoder_msmarco_bert-base_german}{\makecell[l]{thuan9889/\\llama\_embedding\_model\_v1}} & 27.37 & 12.91 & 14.74 & 37.49 & 45.83 & 26.86 & 27.53 \\
\hline
\href{https://huggingface.co/PM-AI/bi-encoder_msmarco_bert-base_german}{\makecell[l]{PM-AI/\\bi-encoder\_msmarco\_bert-base\_german}} & 34.21 & 20.30 & 20.79 & 46.67 & 53.92 & 31.42 & 34.55 \\
\hline
\href{https://huggingface.co/sentence-transformers/msmarco-distilbert-multilingual-en-de-v2-tmp-lng-aligned}{\makecell[l]{sentence-transformers/\\msmarco-distilbert-multilingual-en-de-v2-\\tmp-lng-aligned}} & 28.95 & 24.43 & 17.96 & 49.12 & 54.91 & 32.73 & 34.68 \\
\hline
\href{https://huggingface.co/sentence-transformers/multi-qa-mpnet-base-cos-v1}{\makecell[l]{
sentence-transformers/\\multi-qa-mpnet-base-cos-v1}} & 30.00 & 20.80 & 18.99 & 51.17 & 58.77 & 31.99 & 35.29 \\
\hline
\href{https://learn.microsoft.com/en-us/azure/ai-services/openai/concepts/models?tabs=python-secure#embeddings-models}{azure-text-embedding-3-large} & 38.42 & 22.25 & 23.39 & 66.68 & 69.30 & 39.13 & 43.20 \\
\hline
\end{tabular}
\caption{An evaluation of the text encoder models to be used with the ensemble of encoders. We used a small, manually-created test collection to assess the capabilities of the available encoders. We selected the top 3 best encoders based on the average across six information retrieval metrics.}
\label{tab:model_selection}
\end{table*}

\subsubsection{Dataset}
\Cref{tab:v2_stats} reports the properties of a small manually created test dataset used to select encoders for the ensemble.

\begin{table}[h!]
\centering
\begin{tabular}{l|l}
\hline
Parameter & Value \\
\hline
\# documents & 79.6K    \\
\# queries  & 20    \\
\# relevant documents & 406    \\
\hline
\end{tabular}
\caption{Small manually created test dataset used to select encoders for the ensemble.}
\label{tab:v2_stats}
\end{table}

\subsubsection{Evaluation and metrics}
\label{sec:metrics}
The models listed in the \Cref{tab:model_selection} were evaluated with multiple information retrieval metrics described below. 

\citet{Liu09} defines the metrics from our evaluation as follows. 

\paragraph{Precision@N} In an information retrieval system that retrieves a ranked list, the top-n documents are the first n in the ranking. Precision at n is the proportion of the relevant top-n documents.

\paragraph{Recall@10} Recall at n is the proportion of the relevant top-n documents given the overall number of relevant documents.

\paragraph{F1@10} is a harmonic mean of precision and recall, providing a single metric that balances the two.

\paragraph{MAP@10} The Mean Average Precision (MAP) is the arithmetic mean of the average precision values for an information retrieval system over a set of n query topics. It can be expressed as follows:

\begin{equation}
MAP@10 = \frac{1}{n} \sum_{n} AP@10_n
\end{equation}

where $AP@N$ represents the Average Precision value for a given topic from the evaluation set of n topics. Average precision is a measure that combines recall and precision for ranked retrieval results. For one information need, the average precision is the mean of the precision scores after each relevant document is retrieved.
\begin{equation}
    AP@10 = \frac{\sum_{r} P@10}{R}
\end{equation}
where $r$ is the rank of each relevant document, $R$ is the total number of relevant documents, and $P@10$ is the precision of the top-10 retrieved documents.

\paragraph{MRR} The Reciprocal Rank (RR) information retrieval measure calculates the reciprocal of the rank at which the first relevant document was retrieved. RR is 1 if a relevant document was retrieved at rank 1; if not, it is 0.5 if retrieved at rank 2, and so on. The measure is called the Mean Reciprocal Rank (MRR) when averaged across queries.

\paragraph{nDCG} Discounted Cumulated Gain (DCG) is an evaluation metric for information retrieval (IR). It is based on non-binary relevance assessments of documents ranked in a retrieval result. It assumes that, for a searcher, highly relevant documents are more valuable than marginally relevant documents. It further assumes that the greater the ranked position of a relevant document (of any relevance grade), the less valuable it is for the searcher because the less likely it is that the searcher will ever examine the document – and at least has to pay more effort to find it. $nDCG$ is a normalized metric calculated on the maximum possible DCG through position p, e.g., 10. 

\subsection{Results}

\Cref{tab:model_selection} reports the evaluation of the selected text encoders. We selected the top 3 best encoders based on the average across six information retrieval metrics, i.e., two public and one commercial model. The commercial model in a multilingual encoder LLM shows a steep metric improvement compared to the public LMs. We assume that having an initial strong encoder in the ensemble can impact the overall result later.

\end{document}